\documentclass[a4paper,twoside]{article}

\usepackage{epsfig}
\usepackage{subfigure}
\usepackage{calc}
\usepackage{amssymb}
\usepackage{amstext}
\usepackage{amsmath}
\usepackage{amsthm}
\usepackage{multicol}
\usepackage{pslatex}
\usepackage{apalike}
\usepackage{SCITEPRESS}
\usepackage[small]{caption}
\usepackage{times,graphicx,subfigure,amsmath,url,balance}

\begin{document}

\title{An Unsupervised Ensemble-based Markov Random Field Approach to Microscope Cell Image Segmentation} 

\author{\authorname{B\'alint Antal, Bence Remenyik, and Andr\'as Hajdu}
\affiliation{University of Debrecen, Faculty of Informatics, POB 12, 4010, Debrecen}
\email{\{antal.balint, hajdu.andras\}@inf.unideb.hu, bencerem@gmail.com}
}


\abstract{In this paper, we propose an approach to the unsupervised segmentation of images using Markov Random Field. The proposed approach is based on the idea of Bit Plane Slicing. We use the planes as initial labellings for an ensemble of segmentations. With pixelwise voting, a robust segmentation approach can be achieved, which we demonstrate on microscope cell images. We tested our approach on a publicly available database, where it proven to be competitive with other methods and manual segmentation.}

\onecolumn \maketitle \normalsize \vfill

\section{Introduction}
\label{sec:introduction}

\begin{figure}[!htb]
	\includegraphics[width=\linewidth]{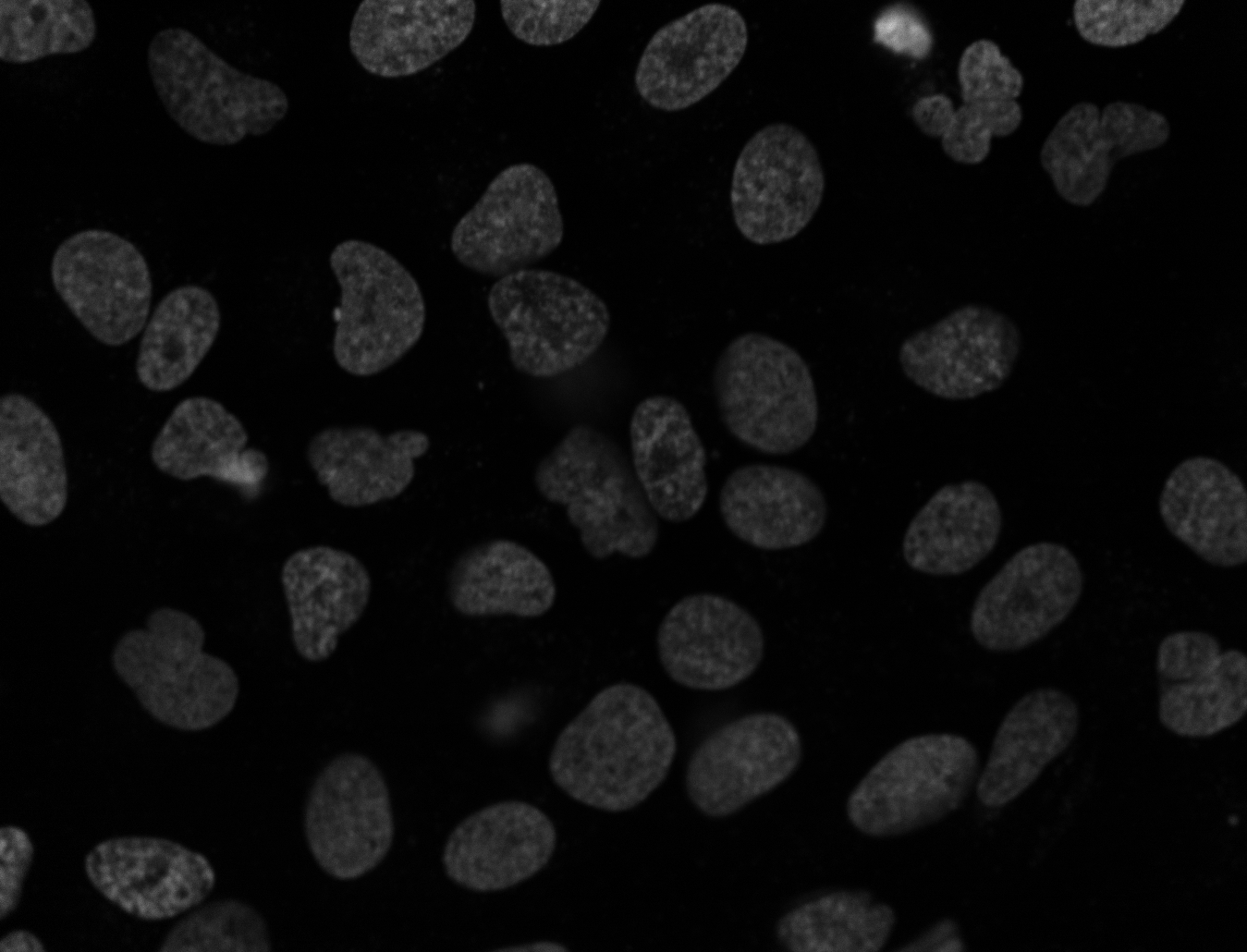}
	\caption{A sample image from the dataset.}
	\label{fig:example}
\end{figure}

Microscope cell segmentation is a very important and challenging task for the medical image processing community as well as physicians. Cell segmentation is essential for several cytometric tasks like cell counting and tracking. The automatic segmentation of cell images is a well-studied field \cite{meijering} \cite{coelho}. However, efficient segmentation of such images is still an open issue. A sample image can be seen in Figure \ref{fig:example}.

In this paper, we present an approach which is shown to be effective in this field. This approach is based on 
Markov Random Field segmentation, which is a very effective way for segmenting images with near-homogeneous objects (like cells). However, the usual way for Markov Random Field segmentation is via supervised learning of certain features, which makes is dependent on the quality of the training data. The proposed method substitutes this weakness with an automatic approach. We provide an automatic initial labelling of the images based on only pixel intensities. Since there are multiple possible choices are available for this task, we run the segmentation from multiple starting points and create an ensemble from them. As the results will demonstrate, our approach outperforms most of the state-of-the-art approaches on a publicly available database and results in a tie with the previous best approaches and the manual segmentation.
  
The rest of the paper is organized as follows: in section \ref{sec:mrf}, we describe the segmentation framework of the Markov Random Fields, which we extend in section \ref{sec:ensemble}. Section \ref{sec:methodology} contains the methodology we used in this study. We present the results in section \ref{sec:results}. Finally, we draw conclusions in section \ref{sec:conclusion}.

\section{Markov Random Field Segmentation}
\label{sec:mrf}

In this section, we briefly summarize the basis for Markov Random Field (MRF) segmentation based on \cite{markov}. Let $I = \{i_{1},\,i_{2},\,\dots,\,i_{n}\}$ be an image. Let $\Lambda = \{0, 1\}$ be a set of labels. Then, we assign each $i_{j}, j=1,\,\dots,\,n$ a label $\omega_{i_{j}}$. Let $X$ be a labelling field. $X$ is a Markov Random Field if $P\left(X = \omega\right)$, for all $\omega \in \Lambda$ and $P\left(\omega_{i_{j}} | \omega_{i_{k}},\, i_{j} \neq i_{k} \right) = P\left(\omega_{i_{j}} | \omega_{i_{k}},\, i_{k} \in N_{i_{j}}\right)$, where $N_{i_{j}}$ is a neighbourhood of $i_{j}$.

The segmentation of an image $I$ with the MRF framework presented above, one must find an optimal labelling. Due to the Hammersley-Clifford Theorem \cite{hct}, we can calculate the global energy for a labelling by summarizing the local energies for each pixels if $P\left(\omega\right)$ follows a Gibbs distribution. We  split the local energy into two terms for all $i_{j}$:
\[
	E_{singleton}\left(i_{j}\right) = P(i_{j} | \omega_{i_{j}} = \dfrac{1}{\sqrt{2\pi}\sigma_{\omega_{i_{j}}}} \exp \left(\dfrac{\left(i_{j} \nu_{\omega_{i_{j}}}  \right)^2}{2\sigma_{\omega_{i_{j}}}}\right),
\]
where $\sigma$ is the standard deviation and the $\nu$ is the mean of the sample.
\[
	E_{doubleton}\left(i_{j}\right) = V\left(j,\,k\right) = \begin{cases}
						- \beta & if \omega_{i_{j}} = \omega_{i_{k}}\\
						\beta & otherwise.
					\end{cases}
\]
The first term considers the distribution of the pixel labels as Gaussian. For this term, the $\sigma$ and $\nu$ must be determined prior segmentation. Usually, this task requires training. The second term is a smoothness prior ensuring homogeneous segmentation of clustered regions. In this case, the global energy $U$ is the following:
\[
	U = \displaystyle\sum_{j = 0}{n} \left(E_{singleton}\left(i_{j}\right) + E_{doubleton}\left(i_{j}\right)\right).
\] 

The optimization of MRF configuration can be done by optimizing $U$. If $P\left(\omega\right)$ follows a Gibbs distribution, simulated annealing \cite{sa} converges to the optimal solution with 1 probability. However, simulated annealing tends to be slow in some cases. However, Iterated Conditional Modes (ICM) \cite{icm} can also be effective if there is a good initial configuration.

\section{Unsupervised MRF-ensembles}
\label{sec:ensemble}

\begin{figure*}[htb]

\centering
\subfigure[Original image]{
\label{fig:orig}
\includegraphics[keepaspectratio,width=0.3\linewidth]{dna-0.png}
}
\quad
\subfigure[Bit plane 0]{
\label{fig:plane0}
\includegraphics[keepaspectratio,width=0.3\linewidth]{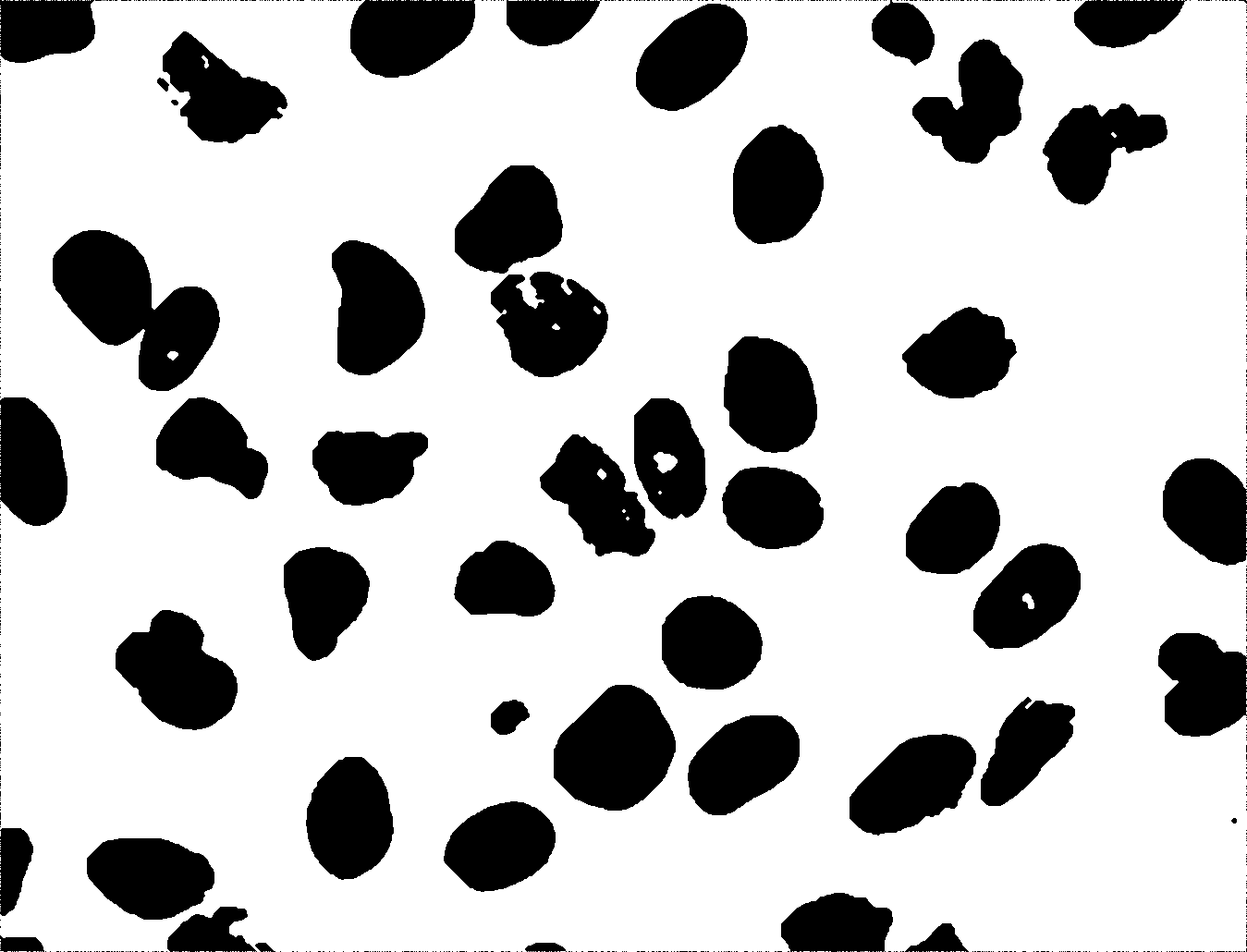}
}
\quad
\subfigure[Bit plane 1]{
\label{fig:plane1}
\includegraphics[keepaspectratio,width=0.3\linewidth]{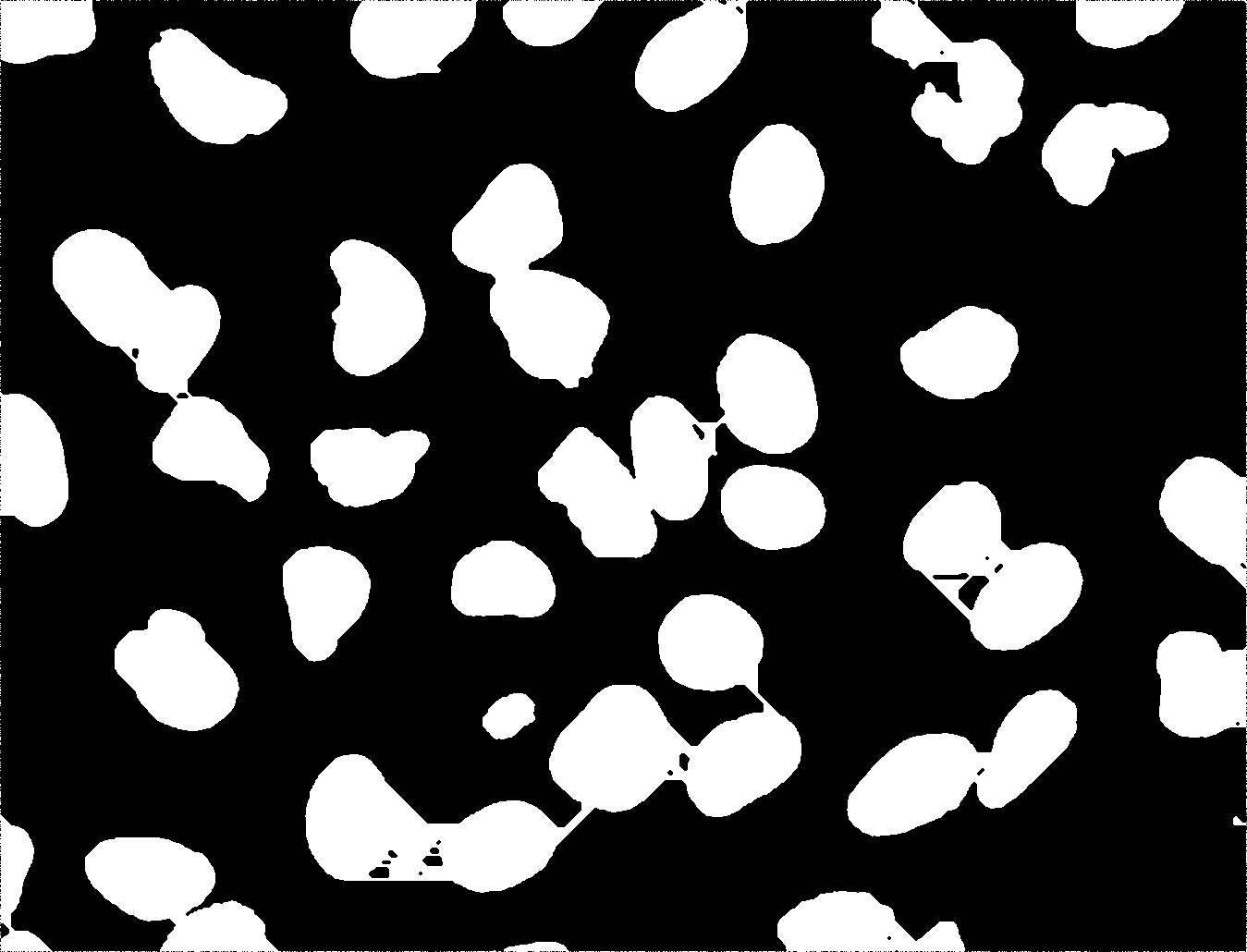}
}
\\
\subfigure[Bit plane 2]{
\label{fig:plane2}
\includegraphics[keepaspectratio,width=0.3\linewidth]{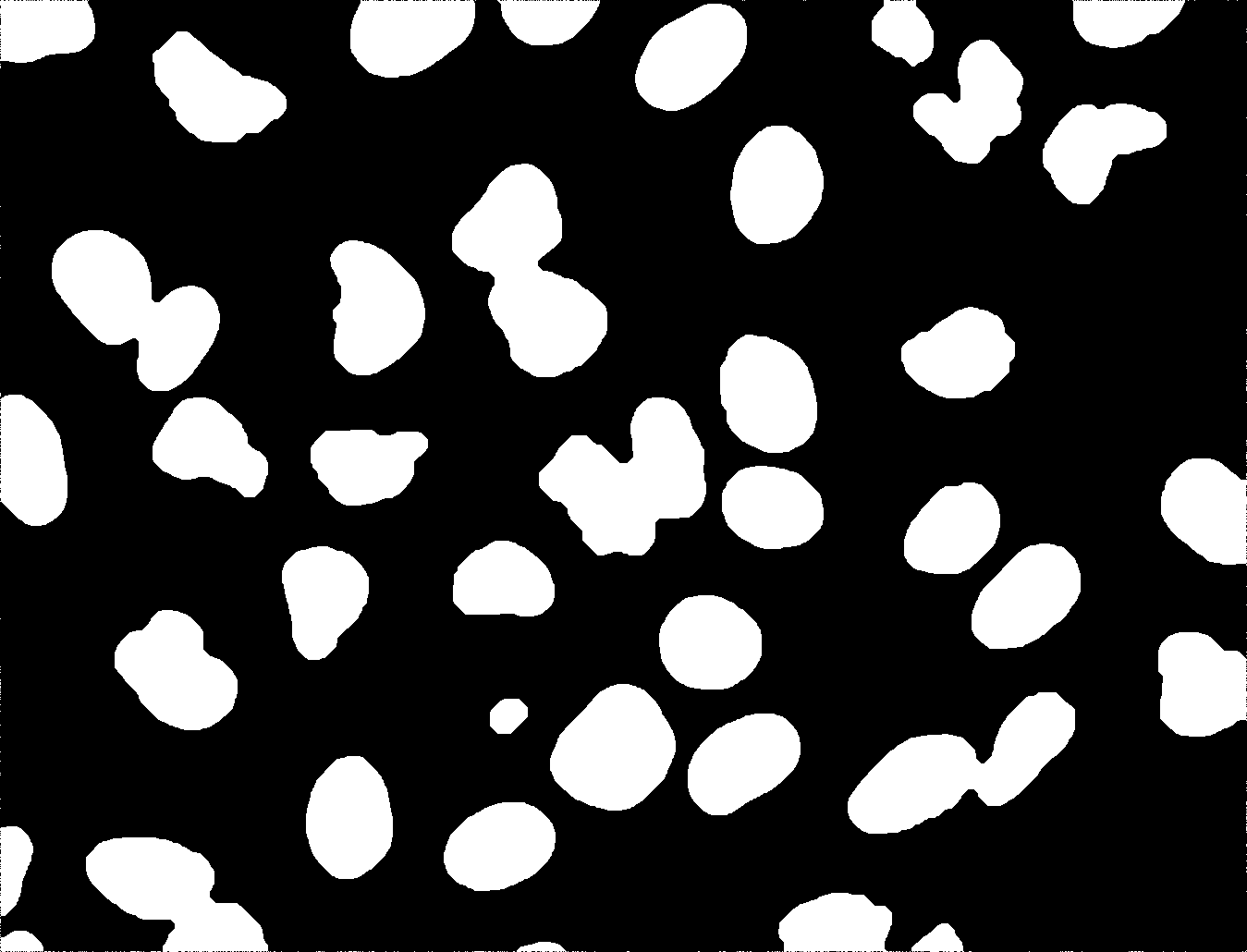}
}
\quad
\subfigure[Bit plane 3]{
\label{fig:plane3}
\includegraphics[keepaspectratio,width=0.3\linewidth]{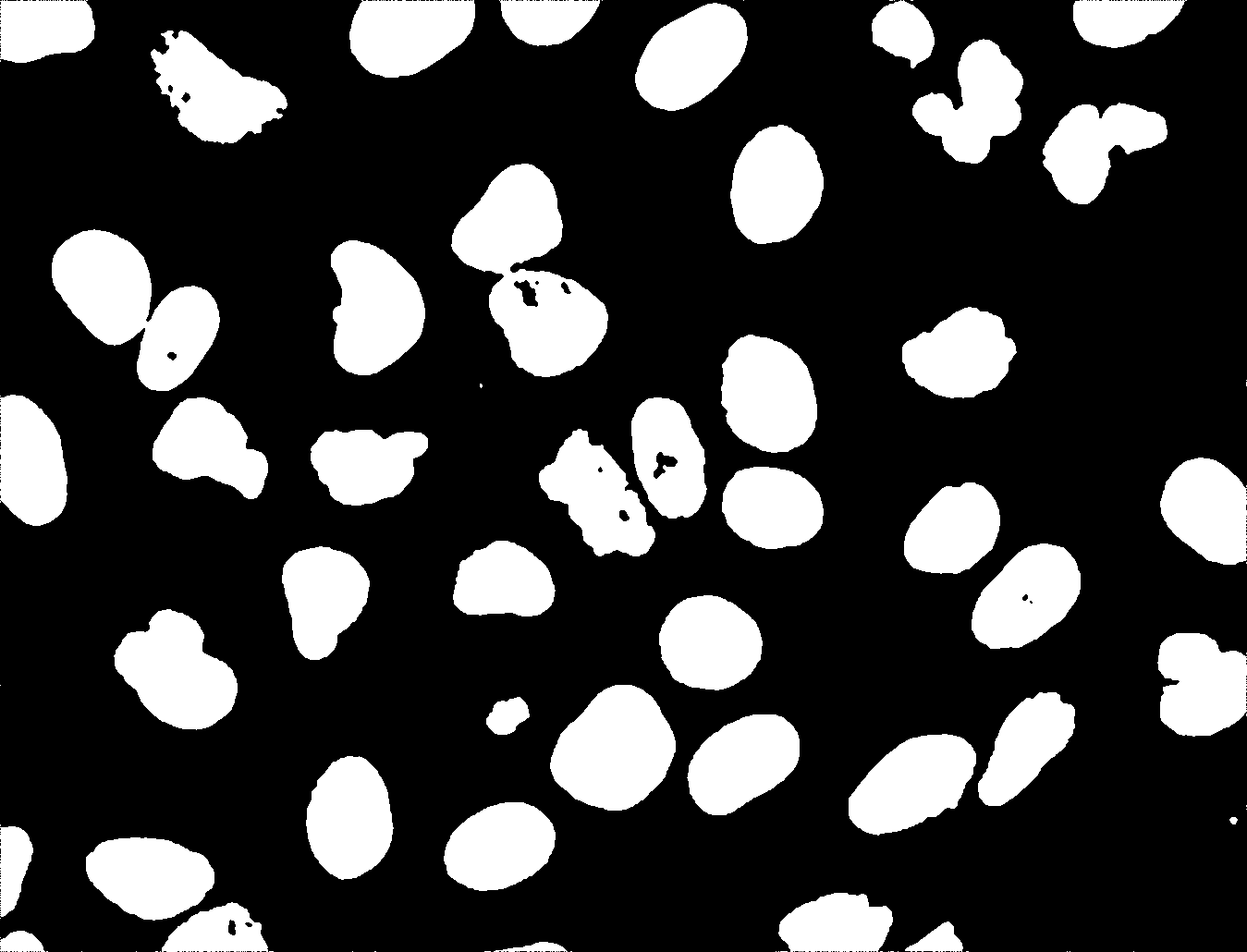}
}
\quad
\subfigure[Bit plane 4]{
\label{fig:plane4}
\includegraphics[keepaspectratio,width=0.3\linewidth]{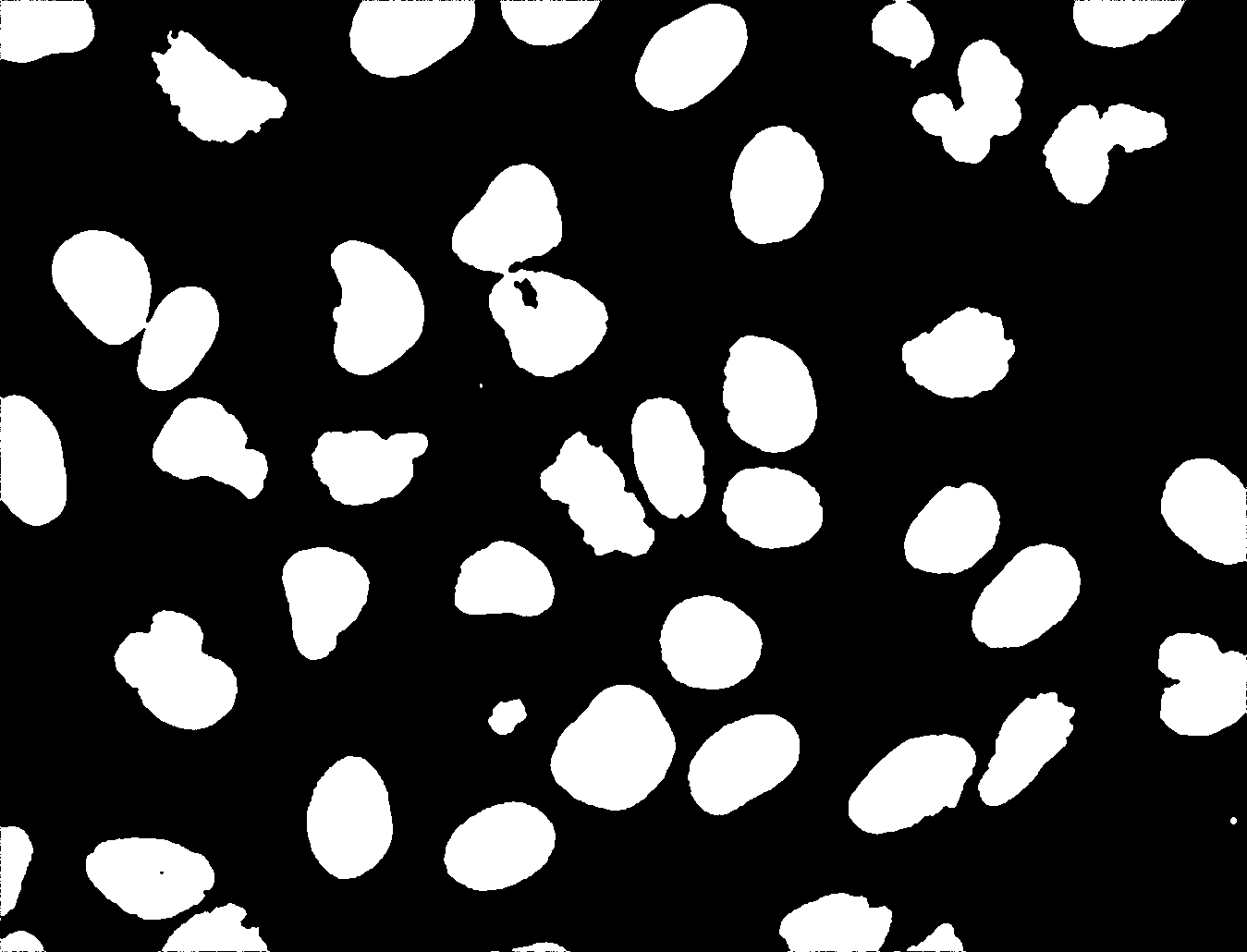}
}
\\
\subfigure[Bit plane 5]{
\label{fig:plane5}
\includegraphics[keepaspectratio,width=0.3\linewidth]{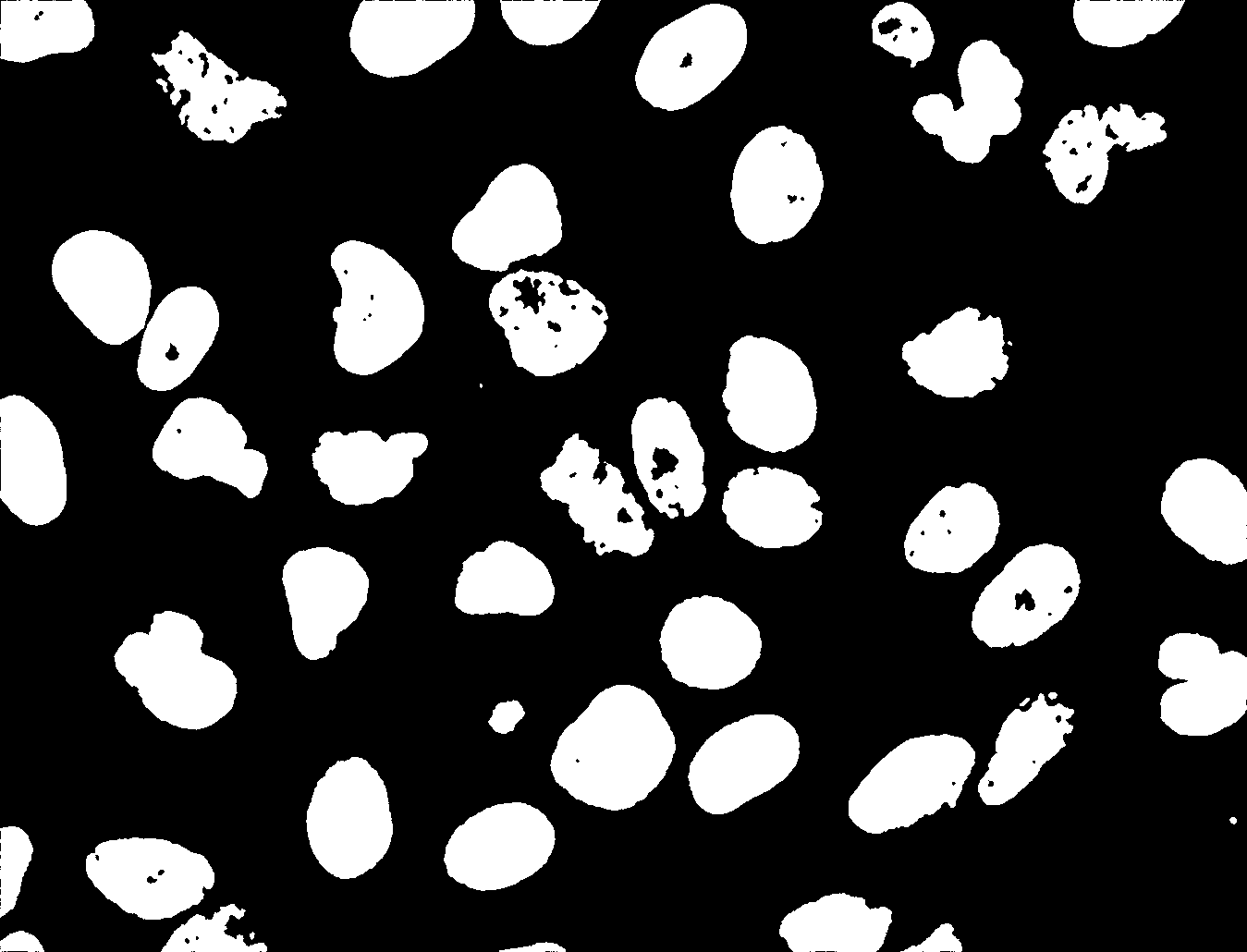}
}
\quad
\subfigure[Bit plane 6]{
\label{fig:plane6}
\includegraphics[keepaspectratio,width=0.3\linewidth]{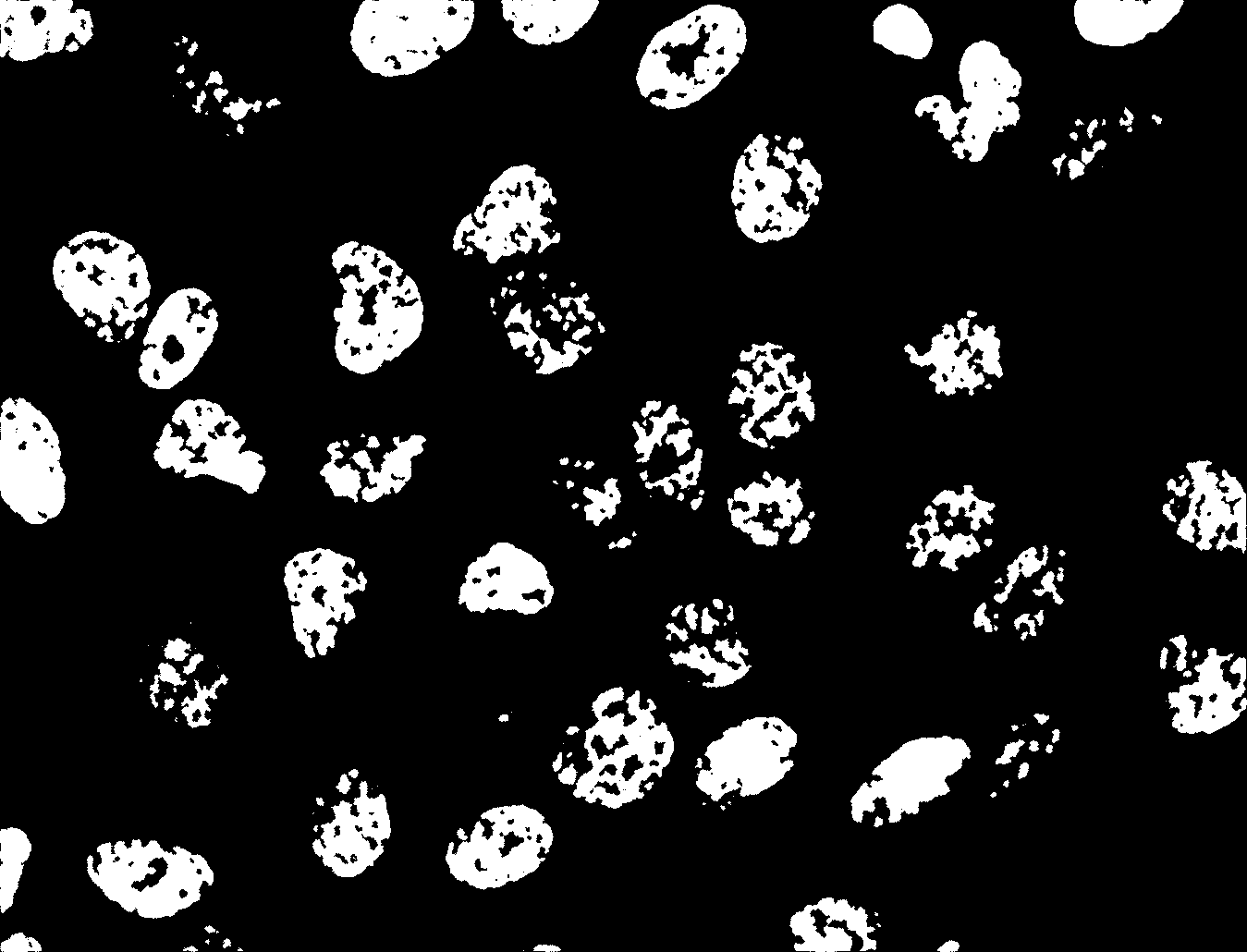}
}
\quad
\subfigure[Bit plane 7]{
\label{fig:plane7}
\includegraphics[keepaspectratio,width=0.3\linewidth]{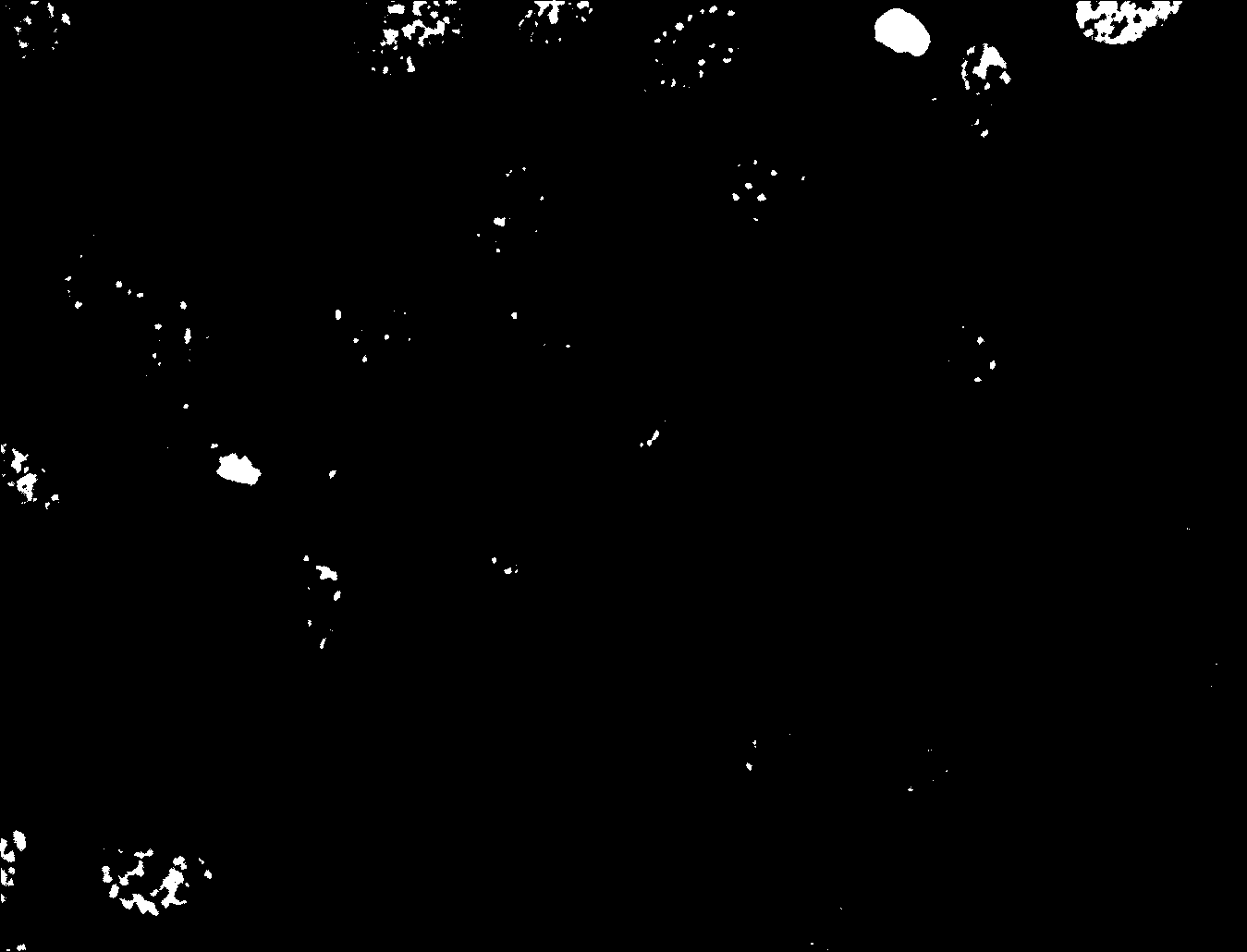}
}
\\
\caption{(a): Original image. (b)-(i) bit planes of \ref{fig:orig}.}
\label{fig:planes}
\end{figure*}

As we stated in Section \ref{sec:mrf}, the usual optimization of MRFs needs training. In this section, we present an approach to lose this dependency. For this task, we use the basic idea of Bit Plane Slicing (BPS) \cite{matlab}. BPS considers an image as a series of planes in the following way:
\[
	BSP\left(j, k\right) = \begin{cases}
						1 & if the jth bit of i_{k} \in I is set\\
						0 & otherwise.
						\end{cases},
\]
where $j = \{0,\,1,\,\dots,\,7$ for a standard 8-bit grayscale image. The planes created by BSP can be seen in Figure \ref{fig:planes} on a sample image. A plane can be regarded as an initial labelling of the original image without having any prior knowledge about the image. In this way, we can calculate the parameters for $E_{singleton}$ and start the optimization process from an initial configuration.

\begin{figure*}[htb]

\centering
\subfigure[Probability map]{
\label{fig:voted}
\includegraphics[keepaspectratio,width=0.3\linewidth]{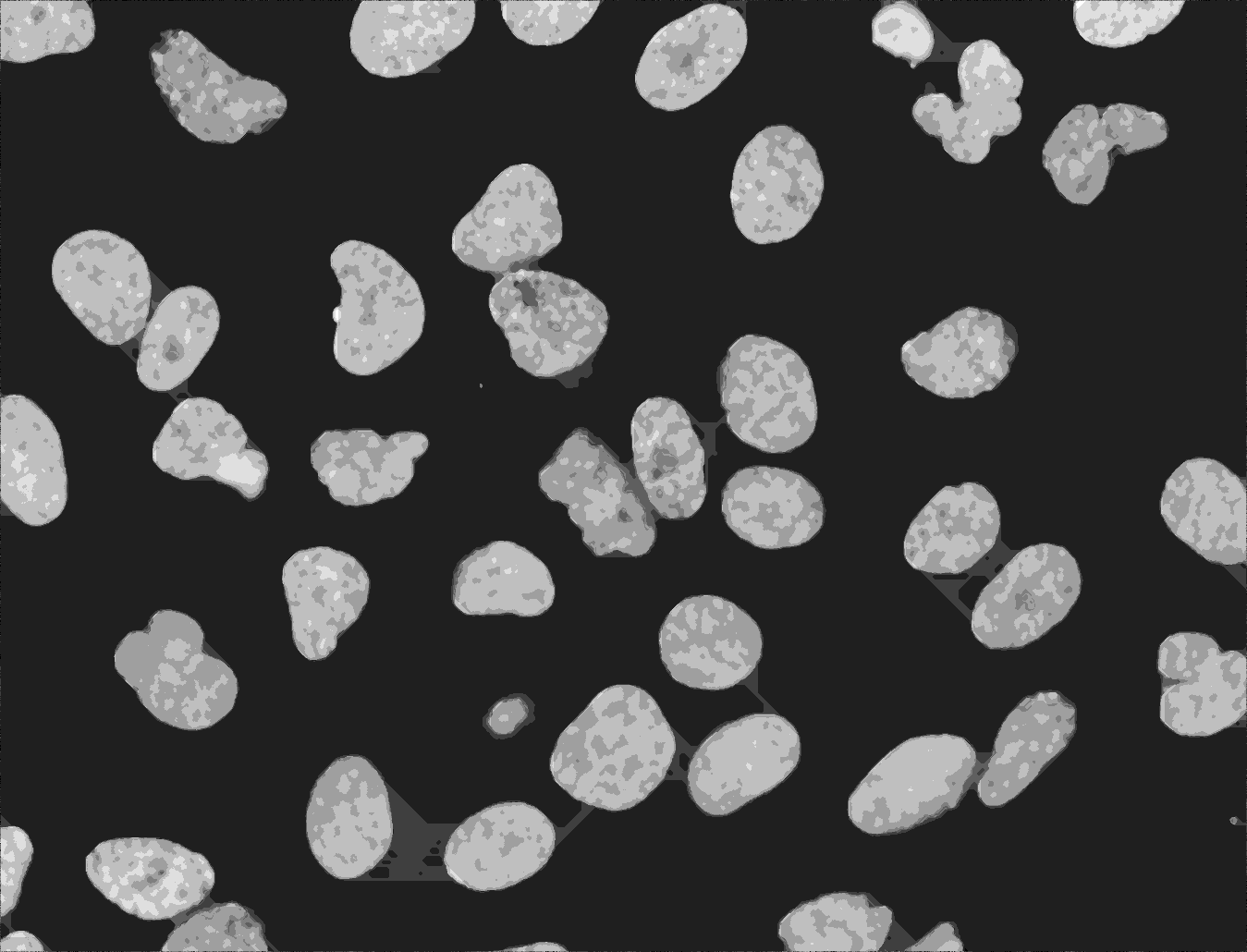}
}
\quad
\subfigure[Confidence level 0]{
\label{fig:cl0}
\includegraphics[keepaspectratio,width=0.3\linewidth]{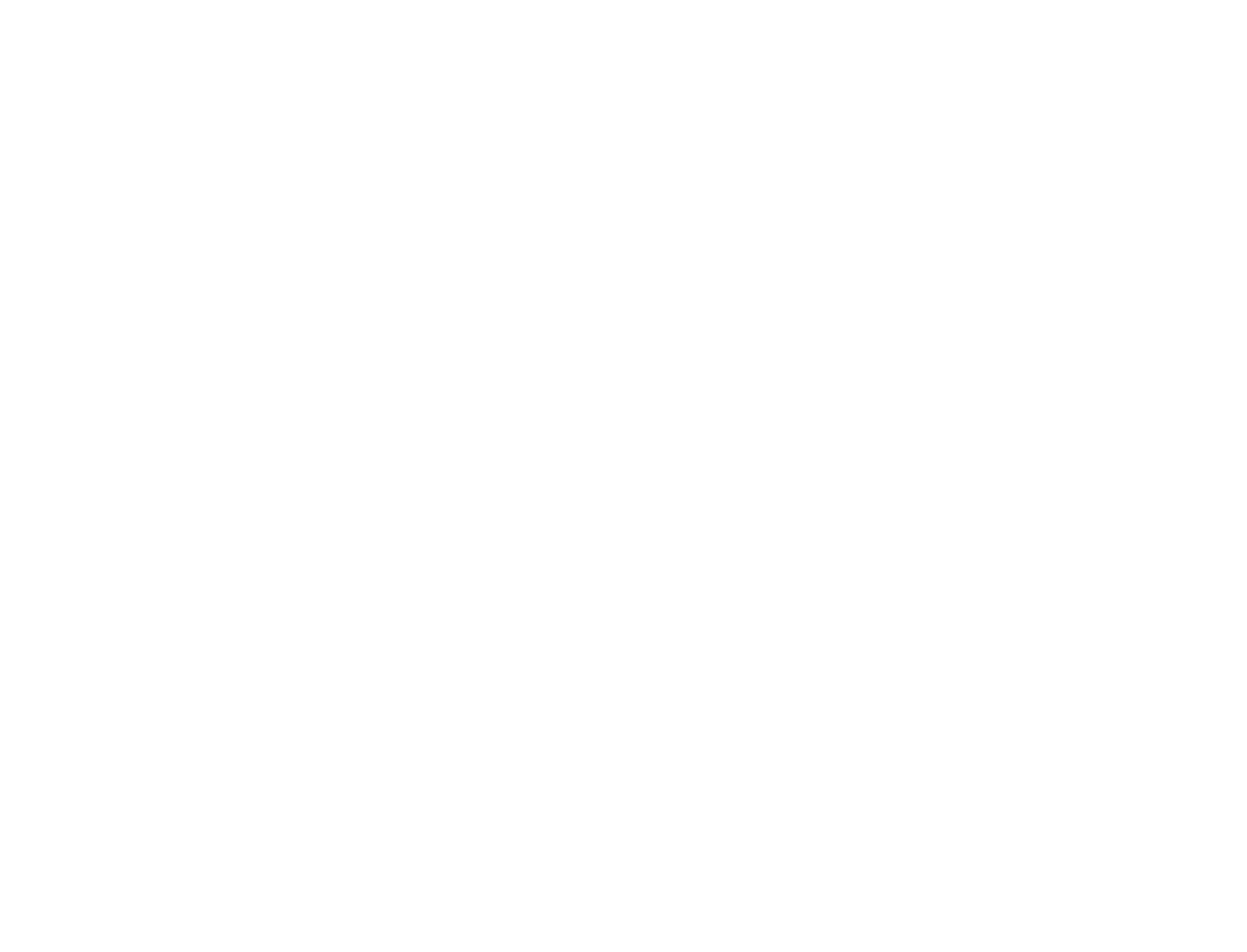}
}
\quad
\subfigure[Confidence level 1]{
\label{fig:cl1}
\includegraphics[keepaspectratio,width=0.3\linewidth]{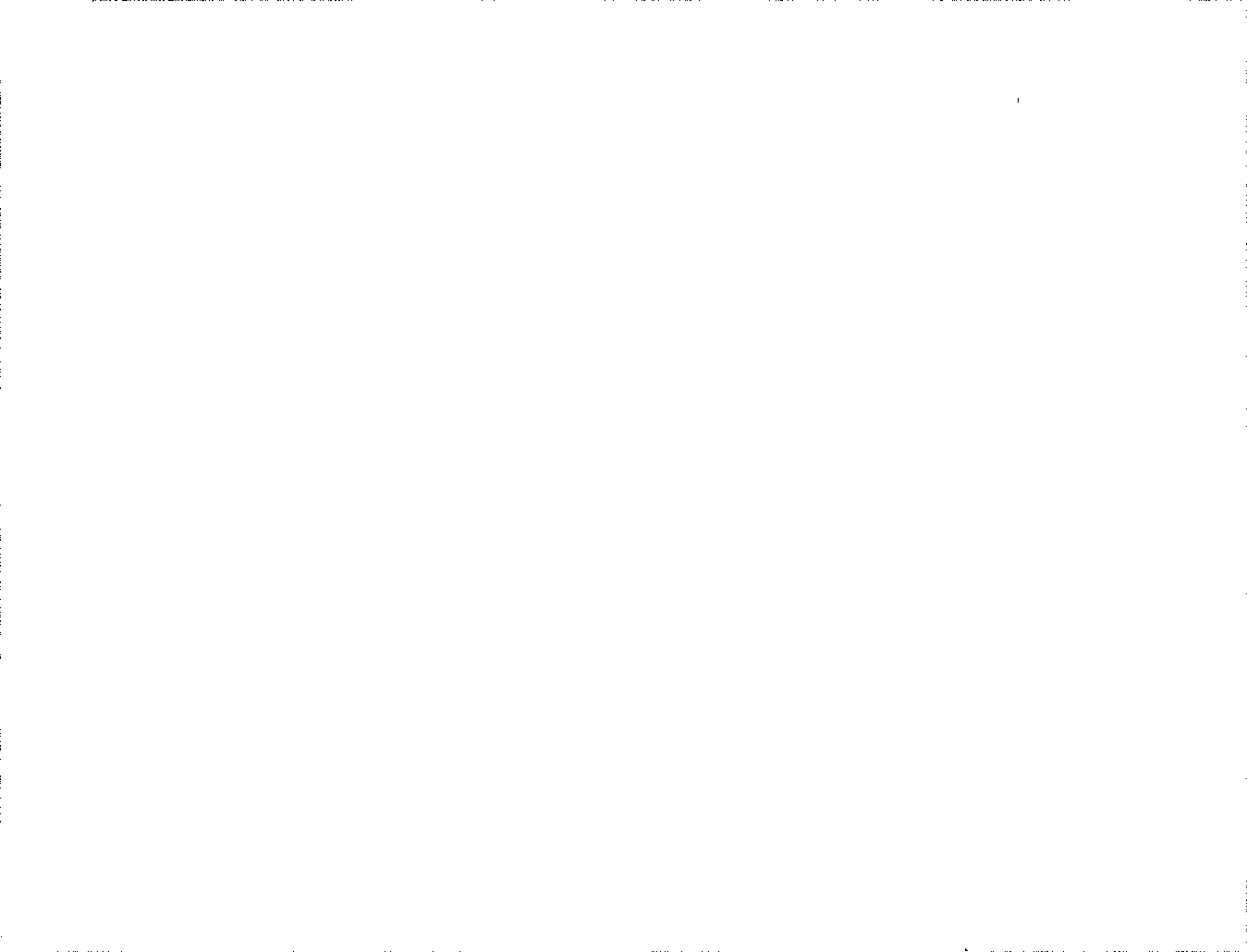}
}
\\
\subfigure[Confidence level 2]{
\label{fig:cl2}
\includegraphics[keepaspectratio,width=0.3\linewidth]{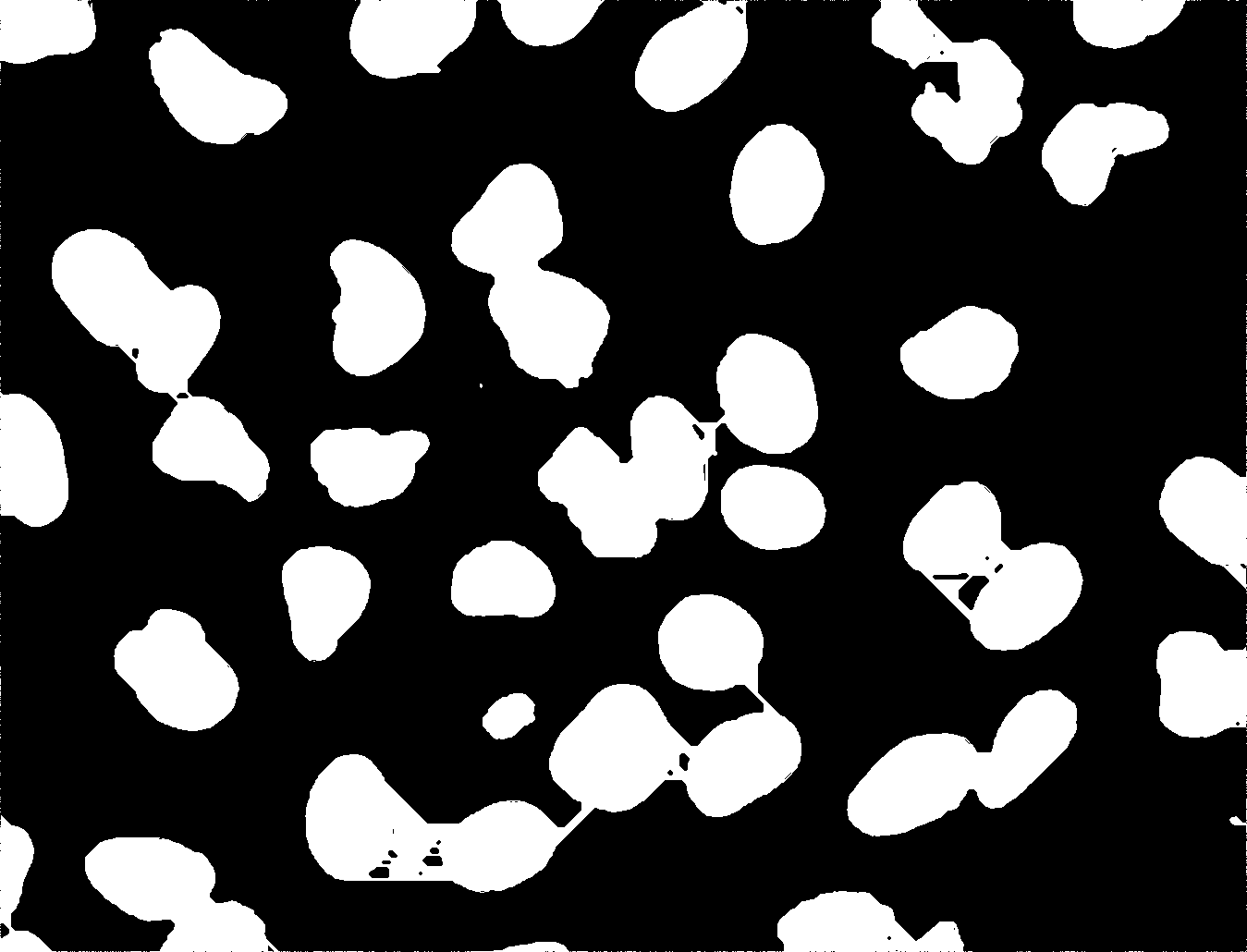}
}
\quad
\subfigure[Confidence level 3]{
\label{fig:cl3}
\includegraphics[keepaspectratio,width=0.3\linewidth]{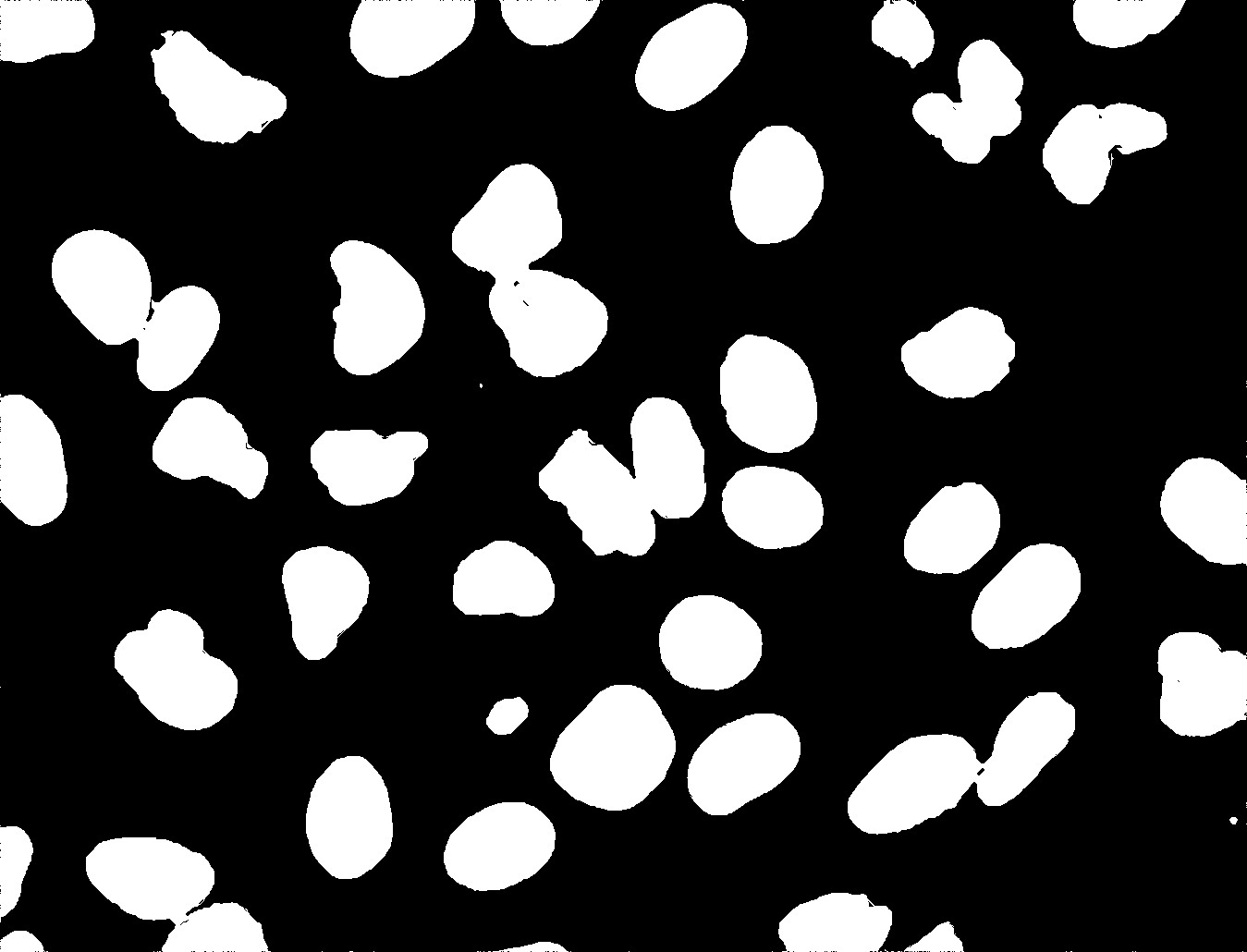}
}
\quad
\subfigure[Confidence level 4]{
\label{fig:cl4}
\includegraphics[keepaspectratio,width=0.3\linewidth]{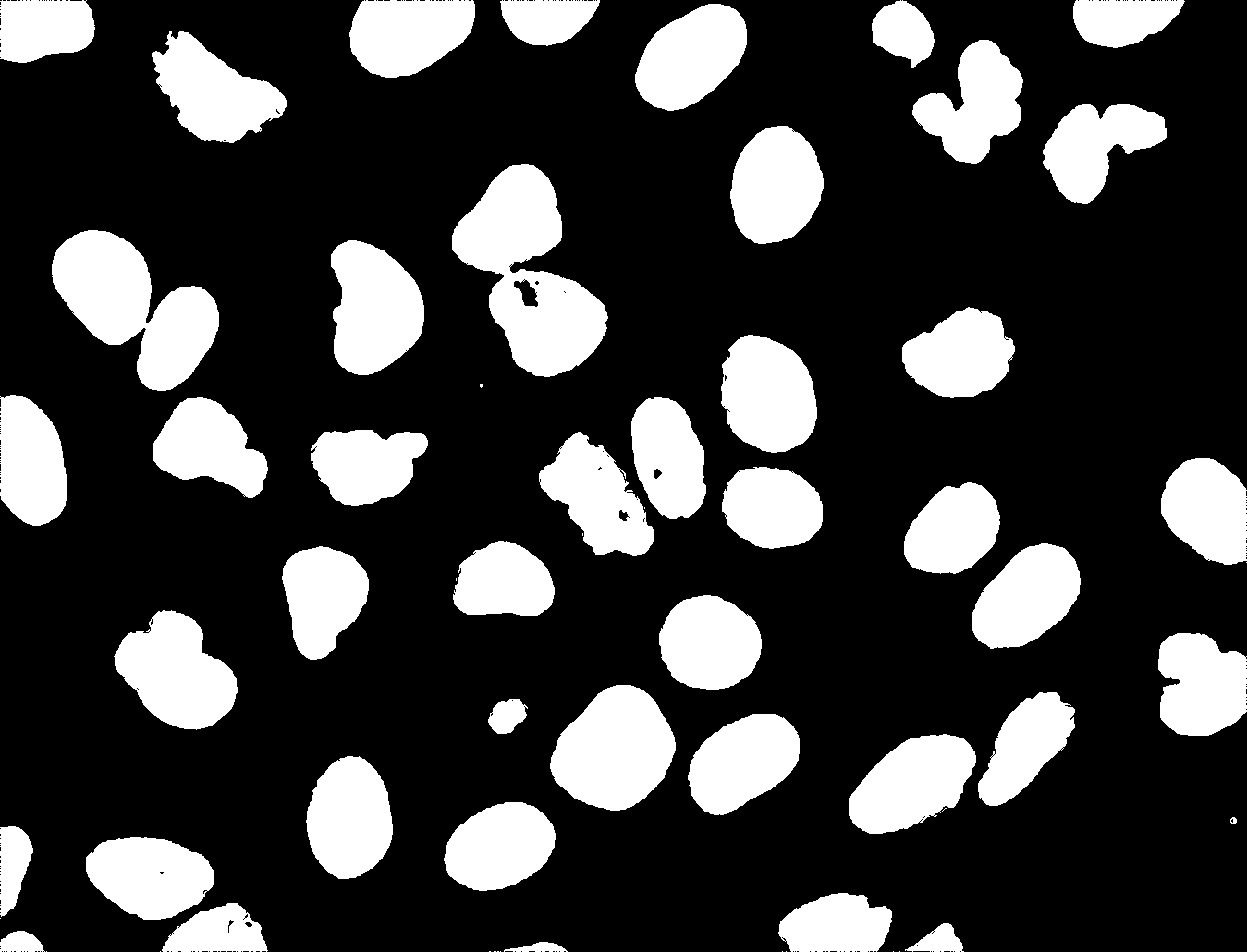}
}
\\
\subfigure[Confidence level  5]{
\label{fig:cl5}
\includegraphics[keepaspectratio,width=0.3\linewidth]{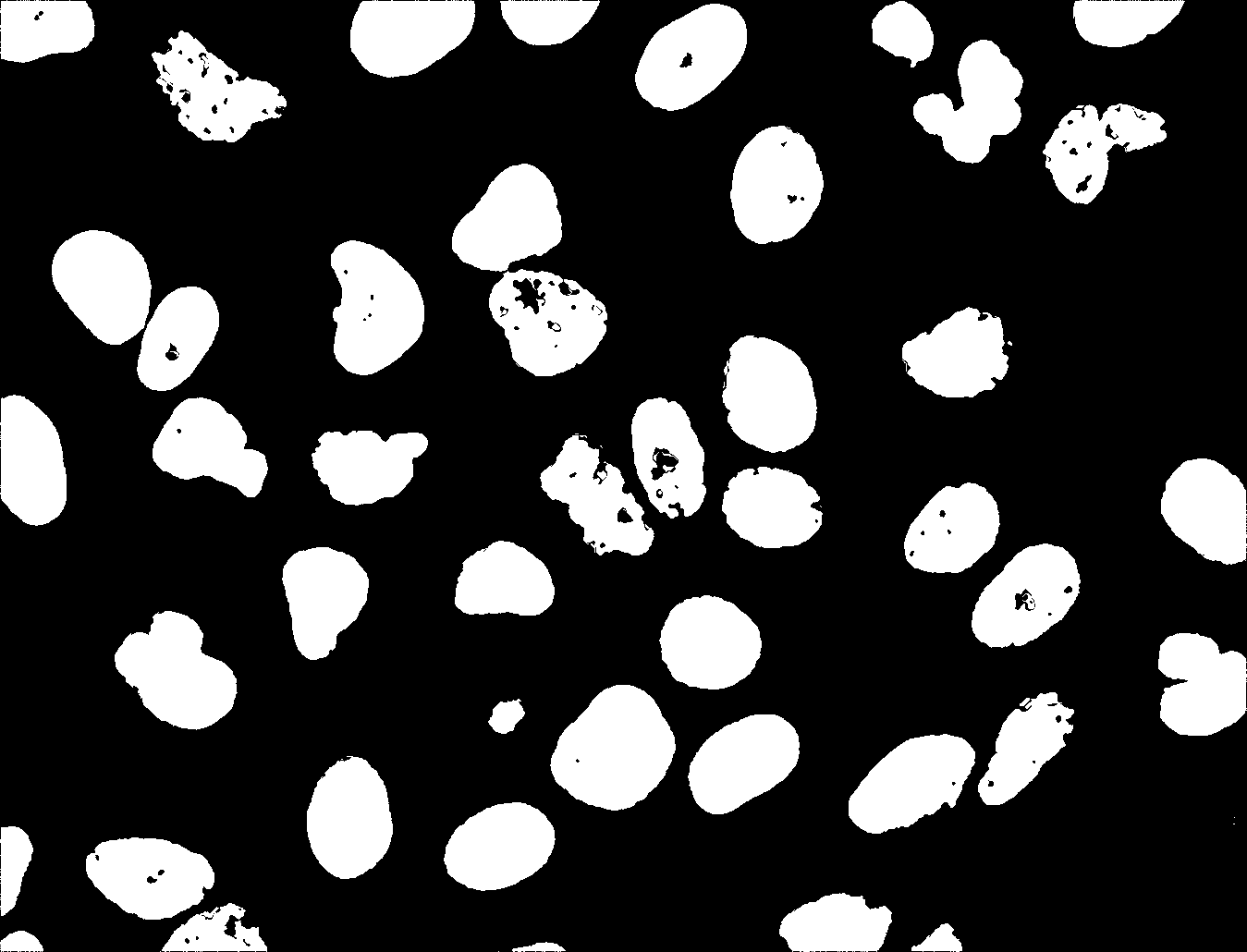}
}
\quad
\subfigure[Confidence level  6]{
\label{fig:cl6}
\includegraphics[keepaspectratio,width=0.3\linewidth]{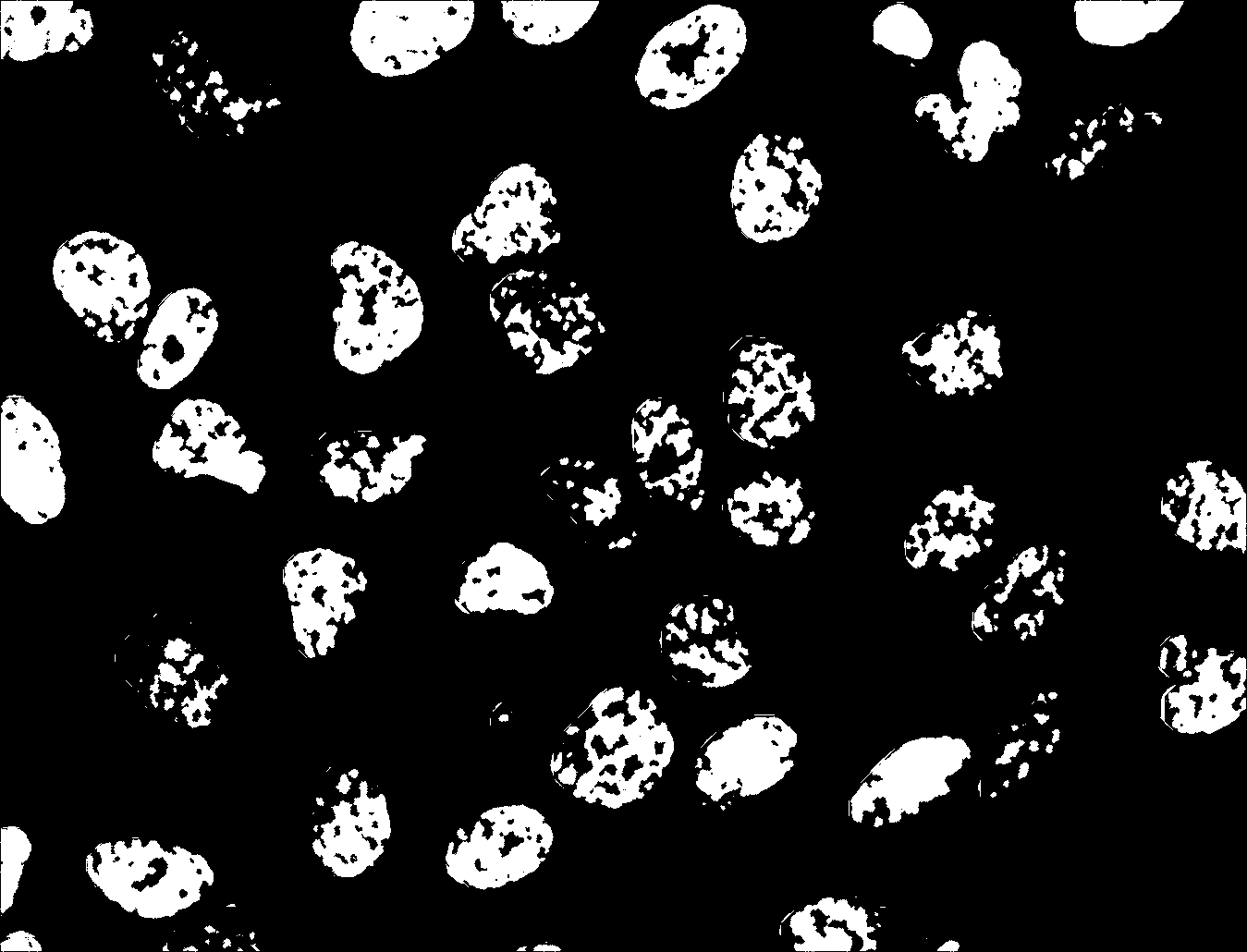}
}
\quad
\subfigure[Confidence level  7]{
\label{fig:cl7}
\includegraphics[keepaspectratio,width=0.3\linewidth]{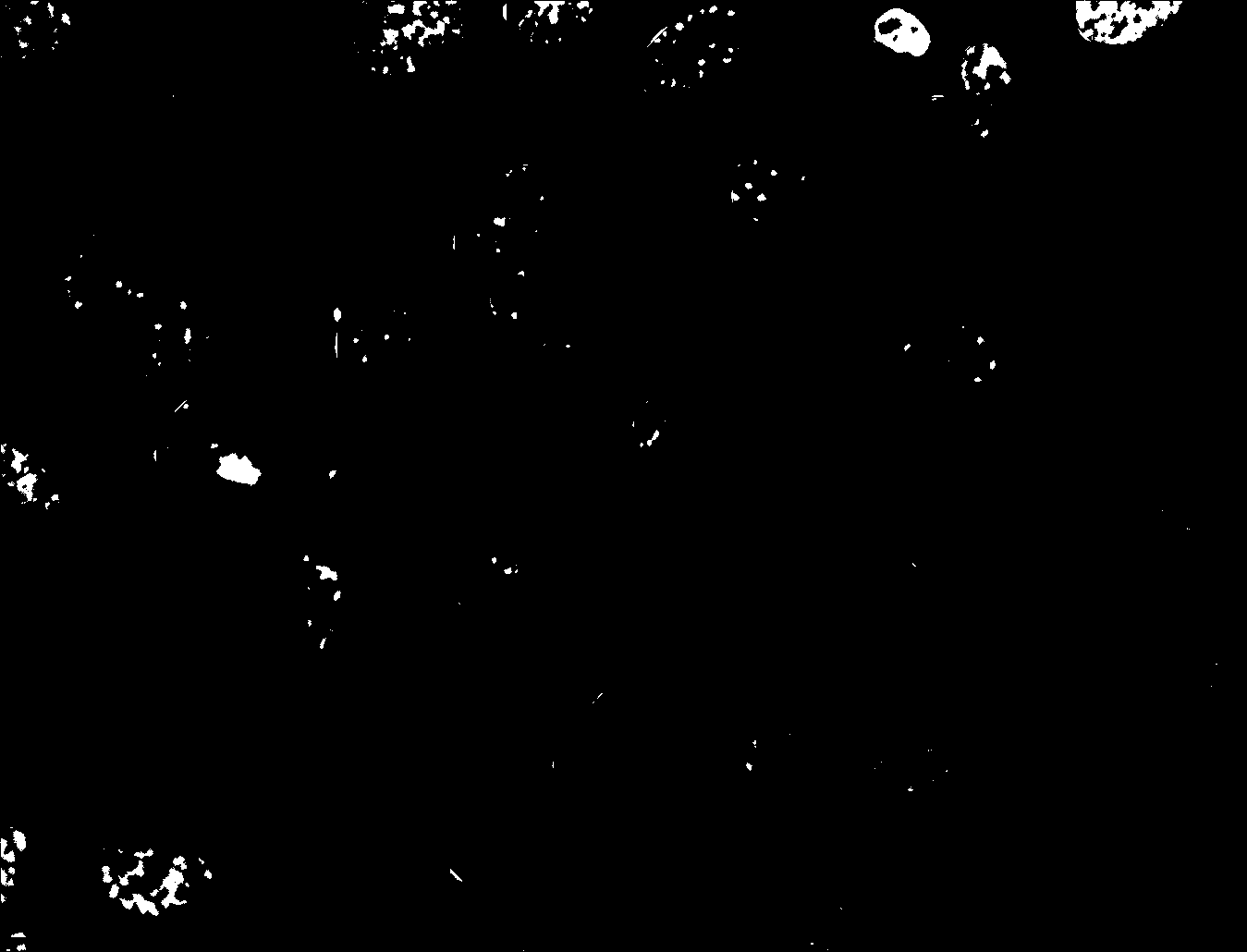}
}
\\
\caption{(a): Probability map for the sample image shown in \ref{fig:orig}. (b)-(i) Voted images with ascending confidence level.}
\label{fig:cls}
\end{figure*}

As no single plane can be selected obviously as a proper initial labelling for an MRF, we propose to use all of them as an ensemble \cite{antal_tbme2012} \cite{antal_pr2012}. That is, we run the optimization eight times using each plane as the initial labelling. Then, we can use pixelwise voting \cite{antal_ispa2011} on the resulting eight images. In this way, each pixel on the resulting image will be having a confidence level between 0 and 7 depending on how many of the segmentations labelled them as object points. In Figure \ref{fig:cls}, we can see a probability map generated from the confidence levels, and the results for thresholding the probability map at the different confidence levels.  

\section{Methodology}
\label{sec:methodology}

In this section, we provide a brief overview on the methodology we used in this experiment. First, in section \ref{sec:database}, we present the database we used. Then, we introduce our evaluation procedure in section \ref{sec:evaluation}. 

\subsection{Database}
\label{sec:database}

We used the U2OS microscope cell image database \cite{coelho}. The database consists of 50 images with 1349 $\times$ 1030 resolution in PNG format. The database contains 1830 cells, which a per image cell count between 24 and 63. We did not use any of the hand-segmented ground truth for learning.

\subsection{Evaluation}
\label{sec:evaluation}

To evaluate our segmentation approach, we have considered several metrics. In this section, we briefly introduce the selected set of evaluation metrics.

For each metrics, we use the following notations. Let $I = \{i_{1},\,i_{2},\,\dots,\,i_{n}\}$ be an image, $S = \{s_{j} \in I\},\, j = 1,\,\dots,\,k,\,k \leq n$ be the result of the segmentation and $G = \{g_{j} \in I\},\, j = 1,\,\dots,\,l,\,l \leq n$ be the ground truth. Then, we use the following notation:
\begin{itemize}
	\item $n_{00} = \displaystyle\sum_{j = 1}^{n} \{1 | i_{j} \notin S \wedge i_{j} \notin G\}.$
	\item $n_{01} = \displaystyle\sum_{j = 1}^{n} \{1 | i_{j} \notin S \wedge i_{j} \in G\}.$
	\item $n_{10} = \displaystyle\sum_{j = 1}^{n} \{1 | i_{j} \in S \wedge i_{j} \notin G\}.$	
	\item $n_{11} = \displaystyle\sum_{j = 1}^{n} \{1 | i_{j} \in S \wedge i_{j} \in G\}.$
	  
\end{itemize}

\subsubsection{Symmetric difference}
\label{sec:sd}

Symmetric difference ($SD$) \cite{mip} is a set theoretic measure counting the elements which belong to either the segmentation or the ground truth bot not both. We also normalize $SD$ with the number of pixels in the image.   That is
\[
	SD = \dfrac{n_{01} + n_{10}}{n}.
\]

\subsubsection{Sensitivity}
\label{sec:sen}

Sensitivity ($SEN$) \cite{kuncheva} is a statistical measure for quantifying the correctly identified positive samples. In our case, it is defined as follows:
\[
	SEN = \dfrac{n_{11}}{n_{11} + n_{01}}.
\]

\subsubsection{Specificity}
\label{sec:spe}

Specificity ($SPE$) \cite{kuncheva} measures the correctly identified negative samples in a binary classification problem. In our case, it is defined as follows:
\[
	SPE = \dfrac{n_{00}}{n_{00} + n_{10}}.
\]

\subsubsection{Positive Predictive Value}
\label{sec:ppv}

Positive Predictive Value ($PPV$) \cite{fscore}  indiciates the proportion of correctly identified positive samples among all samples marked as object points:
\[
	PPV = \dfrac{n_{11}}{n_{11} + n_{10}}.
\]

\subsubsection{F-score}
\label{sec:fscore}

F-score ($SPE$) \cite{fscore} indiciates the proportion of correctly identified positive samples among all samples marked as object points:
\[
	FSCORE = \dfrac{2 \cdot SEN \cdot PPV}{SEN + PPV} .
\]

\subsubsection{Rand Index}
\label{sec:ri}

Rand Index ($RI$) \cite{rand} measure the agreement between the segmentation and the ground truth in the following way:
\[
	FSCORE = \dfrac{n_{11} + n_{00}}{n_{11} + n_{00} + n_{01} + n_{10}} .
\]

\subsubsection{Receiver Operating Characteristics}

We also disclose the Receiver Operating Characteristics (ROC) \cite{likelihood} curve for our segmentation approach. For the curve fitting and for the ROC-related calculations, we used JROCFIT \cite{jrocfit}.

\section{Results}
\label{sec:results}

In Table \ref{tab:confidence}, we can see the different evaluation metric values at the different confidence levels. As we can see, the proposed segmentation approach performs best at the 3 confidence threshold. In this way, a sensitivity of 0.84 and a specificity of 0.99 can be achieved.  

\begin{table*}
\centering
\begin{tabular}{|c|c|c|c|c|c|c|c|c|}
\hline
\textbf{Confidence level}	& \textbf{0} & \textbf{1} & \textbf{2} & \textbf{3} & \textbf{4} & \textbf{5} & \textbf{6} & \textbf{7}\\
\hline
$SD$&144.91&107.64&21.71&2.34&2.75&4.09&9.72&14.35\\
\hline
$SEN$&1.00&0.97&0.88&0.84&0.81&0.71&0.39&0.14\\
\hline
$SPE$&0.00&0.24&0.84&0.99&0.99&1.00&1.00&1.00\\
\hline
$PPV$&0.26&0.31&0.66&0.98&0.98&0.98&0.98&0.97\\
\hline
$FSCORE$&0.41&0.47&0.75&0.90&0.89&0.83&0.55&0.24\\
\hline
$RI$&0.26&0.43&0.85&0.96&0.95&0.92&0.84&0.78\\
\hline
\end{tabular}
\caption{Detailed results for the proposed method.}
\label{tab:confidence}
\end{table*}

\begin{figure}
	\includegraphics[width=\linewidth]{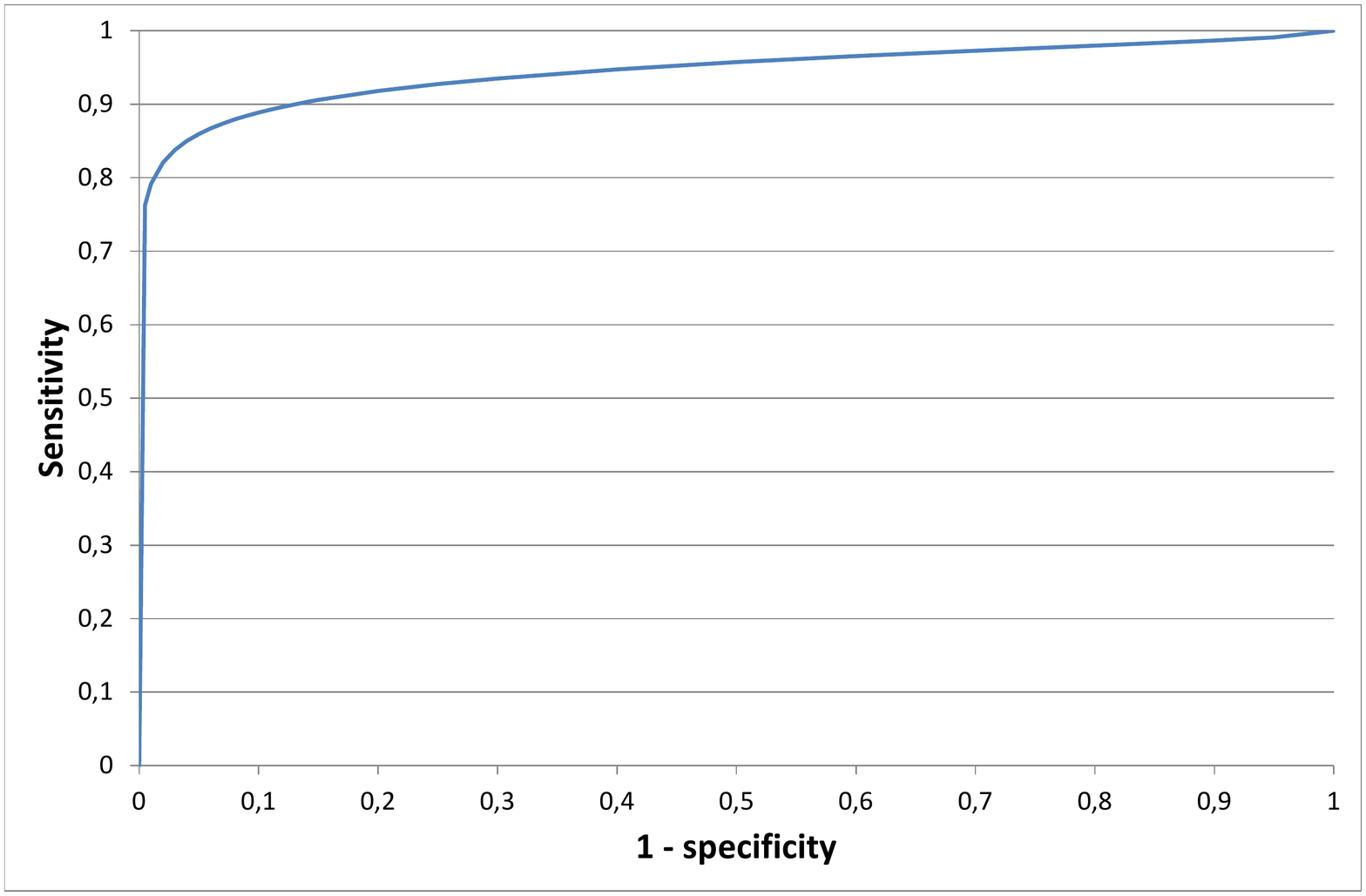}
	\caption{Receiver Operating Characteristics curve for the proposed approach.}
	\label{fig:roc}
\end{figure}

We also evaluated the overall performance of the proposed segmentation approach. The Receiver Operating Characteristics (ROC) curve of the approach can be seen in Figure \ref{fig:roc}. The area under the fitted ROC is 0.945, which indicates a good overall performance on the U2OS database.

We have compared the Rand Index value achieved by our approach to other published results on this database. In all cases, we considered the values presented in \cite{coelho}. As it can be seen, our approach is competitive with the other methods (outperforming five of them) as well as the manual segmentation of an expert.  

\begin{table}
\centering
\begin{tabular}{|c|c|}
\hline
\textbf{Approach} & \textbf{RI}\\
\hline
proposed & 0.96\\
\hline
Mean Threshold & 0.96\\
\hline
Merging Algorithm \cite{lin} & 0.96\\
\hline
AS Manual & 0.96\\
\hline
RC Threshold \cite{ridler} & 0.92\\
\hline
Otsu Threshold \cite{otsu} & 0.92\\
\hline
Watershed (direct) & 0.91\\
\hline
Watershed (gradient) & 0.90\\
\hline
Active Masks \cite{srinivasa} & 0.87\\
\hline
\end{tabular}
\caption{Comparison of the proposed method with other approaches.}
\label{tab:comparison}
\end{table}  

\section{Conclusion}
\label{sec:conclusion}

In this paper, we presented an approach to the  unsupervised segmentation of images using Markov Random Field. In this way, we can benefit from the well-studied and efficient framework of MRFs without the dependency on training. We have demonstrated our approach on the problem of microscope image segmentation, where it performed competitively with other approaches on a publicly available database. In the future, we plan to extend this method to cell tracking on videos.
 
\balance

\vfill
\bibliographystyle{apalike}
{\small
\bibliography{latex8}}

\section*{Acknowledgement}
The publication was supported by the T\'AMOP-4.2.2.C-11/1/KONV-2012-0001 project. The project has been supported by the European Union, co-financed by the European Social Fund.

\vfill
\end{document}